%% file: 2016-cvpr-dynamicSfM.tex
\documentclass[letterpaper, 10 pt, conference]{ieeeconf}
\IEEEoverridecommandlockouts  

\let\labelindent\relax

\input{preamble/commonNoTikz}

\newcommand{\tS}{\tilde{S}}
\newcommand{\tM}{\tilde{M}}
\newcommand{\Kproj}{K_{\textrm{proj}}}
\newcommand{\Ksym}{K_{\textrm{symil}}}
\newcommand{\Kcent}{K_{\textrm{center}}}
\newcommand{\Kupg}{K_{\textrm{upg}}}

\title{\huge{A Factorization Approach to\\ Inertial Affine Structure from Motion}}

\author{%
Roberto Tron%
\thanks{R. Tron is with the Department of Mechanical Engineering at Boston University, Boston, MA 02215, USA \texttt{tron@bu.edu}}%
}

\begin{document}

\maketitle

\begin{abstract}
We consider the problem of reconstructing a {3-D} scene from a moving camera with high frame rate using the affine projection model. This problem is traditionally known as \emph{Affine Structure from Motion} (Affine SfM), and can be solved using an elegant low-rank factorization formulation. In this paper, we assume that an accelerometer and gyro are rigidly mounted with the camera, so that synchronized linear acceleration and angular velocity measurements are available together with the image measurements. We extend the standard Affine SfM algorithm to integrate these measurements through the use of image derivatives.
\end{abstract}

\section{Introduction}
A central problem in geometry in computer vision is \emph{Structure from Motion} (SfM), which is the problem of reconstructing a 3-D scene from sparse feature points tracked in the images of a moving camera. This problem is known also in the robotics community as \emph{Simultaneous Localization and Mapping} (SLAM). One of the main differences between the two communities is that in SLAM it is customary to assume the presence of an \emph{Inertial Measurement Unit} (IMU) that provides measurements of angular velocity and linear acceleration in the camera's frame. Conversely, in SfM there is a line of work which uses an \emph{affine} camera model, which is an approximation to the projective model when the depth of the scene is relatively small with respect to the distance between camera and scene. The resulting \emph{Affine SfM} problem affords a very elegant closed-form solution based on matrix factorization and other linear algebra operations \cite{Tomasi:IJCV92}. This solution has not been used in the robotics community, possibly due to the fact that it cannot be immediately extended to use IMU measurements. 

We assume that the relative pose between IMU and camera has been calibrated using one of the existing offline \cite{Lobo:IJRR07}, online \cite{Jones:IJRR11,Jones:WDV07,Lynen:IROS13,Weiss:ICRA12,Kelly:IJRR11} or closed form \cite{DongSi:IROS12,Martinelli:FTROB13,Martinelli:TRO12} approaches.

\paragraph*{Paper contributions} In this paper we bridge the gap between the two communities by proposing a new \emph{Dynamic Affine SfM} technique. Our technique is a direct extension of the traditional Affine SfM algorithm, but incorporates synchronized IMU measurements. This is achieved by assuming that the frame rate of the camera is high enough and that we can compute the higher order derivatives of the point trajectories. Remarkably, our formulation leads again to a closed form solution based on matrix factorization and linear algebra operations. To the best of our knowledge, this kind of relation between higher-order derivatives of image trajectories (flow) and IMU measurements, and the low-rank factorization relation between them, have never been exploited before.

 
\input{priorWork}

\input{notation}
\input{solution}

\section{Preliminary results} 
\begin{figure*}
  \centering
  \hfill\subfloat[Top view]{\includegraphics{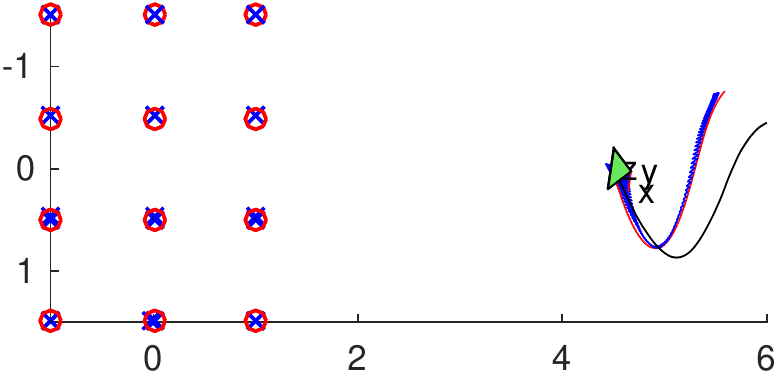}}\hfill
  \subfloat[Front view]{\includegraphics{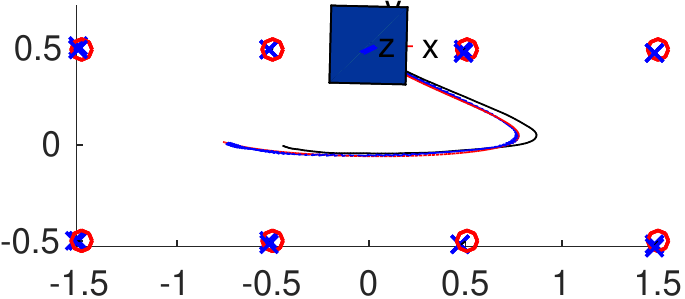}}\hfill
\vspace{-1mm}
  \caption{Simulation results for a preliminary implementation of the Dynamic Affine SfM reconstruction under significantly noisy conditions. Red: ground-truth structure and motion. Blue: reconstructed structure and motion. Black line: motion estimate from integration of IMU measurements alone. Green pyramid: initial reconstructed camera pose. The camera rotates up to $\unit[30]{deg}$ and reaches velocities of up to $\unitfrac[0.5]{m}{s}$. All axes are in meters.
}
  \label{fig:result-affine}
\end{figure*}

\begin{figure*}
  \centering
  \includegraphics{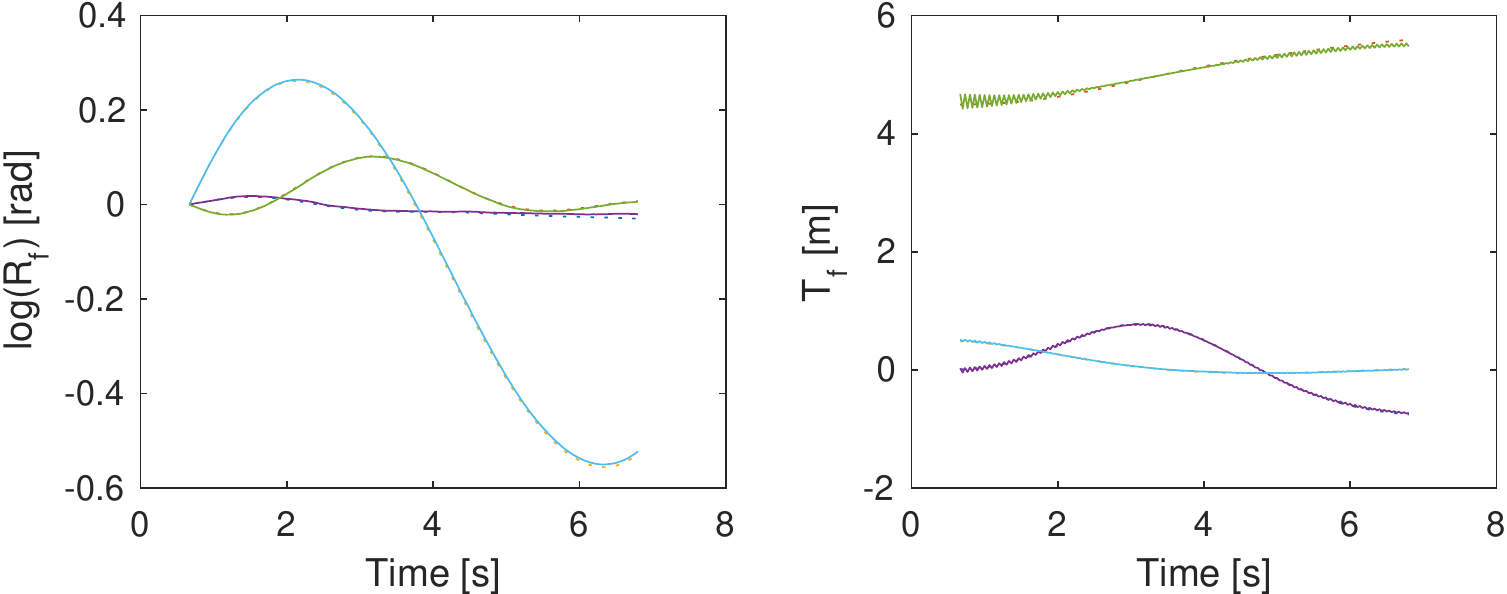}
  \caption{Plots of the ground-truth and estimated trajectories. Left: Absolute rotations in exponential coordinates from the identity, $\log(R_f)$. Right: absolute translations.}
  \label{fig:result-affine-trajectory}
\end{figure*}

Figure \ref{fig:result-affine} shows a simulation of the result of a preliminary implementation of the Dynamic Affine SfM procedure. We have simulated 5 seconds of a quadrotor following a smooth trajectory while an onboard camera tracks 24 points. The measurements (point coordinates, angular velocity and linear acceleration) are sampled at $30\mathrm{Hz}$ and corrupted with Gaussian noise with variances of the added noise: $\unitfrac[3]{deg}{s}$ angular velocity, $\unitfrac[0.2]{m}{s^2}$ acceleration, $\unit[0.5]{\%}$ image points (corresponding to, for instance, $\unit[3.2]{px}$ on a $\unit[600\times600]{px}$ image). The reconstruction obtained using our implementation is aligned to the ground-truth using a Procrustes procedure without scaling and compared with an integration of the inertial measurements alone. Figure \ref{fig:result-affine-trajectory} compares the plot of the ground truth and estimated rotations and translations in absolute coordinates.
Three facts should be noted in this simulation:
\begin{enumerate*}
\item The use of images greatly improves the accuracy with respect to the use of IMU measurements alone.
\item The noise in the estimation mostly appears along the $z$-axis direction of the camera, for which the images do not provide any information.  Although ours is a preliminary implementation, the result obtained is extremely close to the ground-truth, except for small errors along the $z$ axis of the camera. These errors are due to the fact that the affine model discards the information along the $z$ axis (the affine model provide little information in this direction, and the reconstruction mostly relies on the noisy accelerometer measurements).
\item Larger noise appears at lower velocities (beginning and end of the trajectory), thus attesting the usefulness of incorporating higher-order derivative information.
\end{enumerate*}

\section{Extensions and future work}
The approach can be easily extended to the case where multiple (non-overlapping) cameras rigidly mounted to the same IMU. In this case, one can constract multiple matrices $W$ (one for each camera), and performs steps \ref{it:factorization}--\ref{it:centering} of our solution independently. The rotations and translations (steps \ref{it:rotations},~\ref{it:translations}) can then be recovered by solving the linear systems \eqref{eq:opt-rotation} and \eqref{eq:opt-translation} joinly over all the cameras by adjusting the corresponding coefficient matrices with the relative camera-IMU poses (which are assumed to be known).
The Dynamic Affine SfM approach can also be potentially extended to the projective camera model by using the approach of \cite{Dai:IJCV13}. Let $\Lambda\in\real{F\times P}$ be a matrix containing all the unknown depths of each point in each view, and let $L=\Lambda\otimes\bmat{1\\1}$. Then, one can find a low-rank matrix $\hat{W}$ by minimizing
\begin{multline}
    \min_{\hat{W},\Lambda,\dot{\Lambda},\ddot{\Lambda}} \norm{\bmat{L\odot W'\\\dot{L}\odot W' + L \odot \dot{W}'\\\ddot{L}\odot W' +2\dot{L}\odot \dot{W}' + L \odot \ddot{W}'}-\hat{W}}^2_F \\+ \mu \norm{\hat{W}}_\ast+f_\Lambda(\Lambda,\dot{\Lambda},\ddot{\Lambda}),\label{eq:opt-projective}
\end{multline}
where $\norm{\cdot}_\ast$ denotes the nuclear norm (which acts as a low-rank prior for $\hat{W}$), $\mu$ is a scalar weight and $f_\Lambda$ relates $\Lambda$ with its derivatives using a derivative interpolation filter. This problem is convex and can be iteratively solved using block coordinate descent (i.e., by minimizing over $\hat{W}$ and the other variables alternatively). The method described in Section~\ref{sc:solution} can then be carried out on the matrix $\hat{W}$ to obtain the reconstruction. Intuitively, \eqref{eq:opt-projective} estimates the projective depths of each point so that we can reduce the problem to the affine case.

We will implement and evaluate these two extensions in our future work.

\bibliographystyle{biblio/ieee}
\bibliography{biblio/IEEEfull,biblio/IEEEConfFull,biblio/OtherFull,%
  biblio/tron,%
  biblio/vision,biblio/visionGeometry,biblio/poseAveraging,biblio/sfm,biblio/bundleAdjustment,%
  biblio/visionTracking,biblio/visionNRSfM,%
  biblio/factorizationAndLowRank,biblio/learning,biblio/subspaceClustering,%
  biblio/control,biblio/systemIdentification,biblio/optimization,%
  biblio/robotics,%
  biblio/slam,biblio/multiSlam,biblio/slamRobust,biblio/slamReduction,biblio/collaborativeLocalization,%
  biblio/misc,biblio/software}

\end{document}

%% file: preamble/commonNoTikz.tex
\input{preamble/fixes}
\input{preamble/pagination}
\input{preamble/graphics}
\input{preamble/math}
\input{preamble/operators}

\input{preamble/utilities}
\input{preamble/notation}


%% file: preamble/fixes.tex
\usepackage{fix-cm}
\usepackage{etex}

\usepackage{dblfloatfix}

\usepackage{nag}


%% file: preamble/pagination.tex
\usepackage{algorithm}
\usepackage{cite}

\usepackage{microtype}

\usepackage{array}
\usepackage{multirow}
\usepackage{booktabs}

\usepackage[inline]{enumitem}

\usepackage{subfig}



%% file: preamble/graphics.tex
\usepackage[table,x11names]{xcolor}

\usepackage{colortbl}

\usepackage{graphicx}


%% file: preamble/math.tex
\usepackage[cmex10]{amsmath}
\usepackage{amssymb,amsfonts}
\makeatletter
\@ifundefined{proof}{%
\usepackage{amsthm}%
}{}
\makeatother



%% file: preamble/operators.tex
\newcommand{\real}[1]{\mathbb{R}^{#1}{}}

\newcommand{\bmat}[1]{\begin{bmatrix}#1\end{bmatrix}}

\newcommand{\transpose}{^\mathrm{T}}

\newcommand{\inverse}{^{-1}}

\DeclareMathOperator{\stack}{stack}

\DeclareMathOperator{\expm}{expm}

\DeclareMathOperator{\conv}{conv}

\newcommand{\norm}[1]{\lVert#1\rVert}

\newcommand{\frob}[1]{\norm{#1}_F}

\newcommand{\vct}[1]{\mathbf{#1}}


%% file: preamble/utilities.tex
\usepackage{comment}

\usepackage{algpseudocode}

\usepackage{pdfcomment}

\usepackage{units}


%% file: preamble/notation.tex
\providecommand{\cR}{\mathcal{R}}


%% file: priorWork.tex
\section{Review of prior work}
In the vision community, the Dynamic SfM problem is related to traditional \emph{Structure from Motion} (SfM), which uses only vision measurements. The standard solution pipeline \cite{Hartley:book04} includes three steps. First, estimate relative poses between pairs of images by using matched features \cite{Lowe:IJCV04,Dong:CVPR15,Bay:CVIU08} and robust fitting techniques \cite{Fischler:CommACM81,Hartley:pami12}.
Second, combine the pairwise estimates either in sequential stages \cite{Snavely:TOG06,Agarwal:COMM11,Agarwal:ECCV10,Frahm:ECCV10,Snavely:CVPR08}, or by using a pose-graph approach \cite{Carlone:ICRA15} (which works only with the poses and not the 3-D structure). Algorithms under the latter category can be divided into \emph{local} methods \cite{Tron:TAC14,Hartley:IJCV13,Aftab:PAMI15,Chatterjee:ICCV13}, which use gradient-based optimization, and \emph{global} methods \cite{Martinec:CVPR07,Arie-Nachimson:3DIMPVT12,Wang:II13}, which involve a relaxation of the constraints on the rotations together with a low-rank approximation. 
The fourth and last step of the pipeline is to use 
Bundle Adjustment (BA) \cite{Triggs:VATP00,Hartley:book04,Engels:PCV06}, where the motion and structure are jointly estimated by minimizing the reprojection error.

In the robotics community, Dynamic SFM is closely related to other Vision-aided Inertial Navigation (VIN) problems. These include: \emph{Visual-Inertial Odometry} (VIO), where only the robots' motion is of interest, and \emph{Simultaneous Localization and Mapping} (SLAM), where the reconstruction (i.e., map) of the environment is also of interest. 
Existing approaches to these problems fall between two extremes. On one end of the spectrum we have \emph{batch} approaches, which are similar to BA with additional terms taking into account the IMU measurements \cite{Bryson:ICRA09,Strelow:IJRR04}. 
If obtaining a map of the environment is not important, the optimization problem can be restricted to the poses alone (as in the pose-graph approach in SfM), using the images and IMU measurements to build a so-called \emph{factor graph} \cite{Indelman:NDRV15,Dellaert:08,Agrawal:IROS06}. To speed-up the computations, some of the nodes can be merged using IMU \emph{pre-integration} \cite{Lupton:TRO12,Carlone:ICRA15}, and \emph{key-frames} \cite{Konolige:TRO08}.

\begin{figure*}
  \centering
  \includegraphics{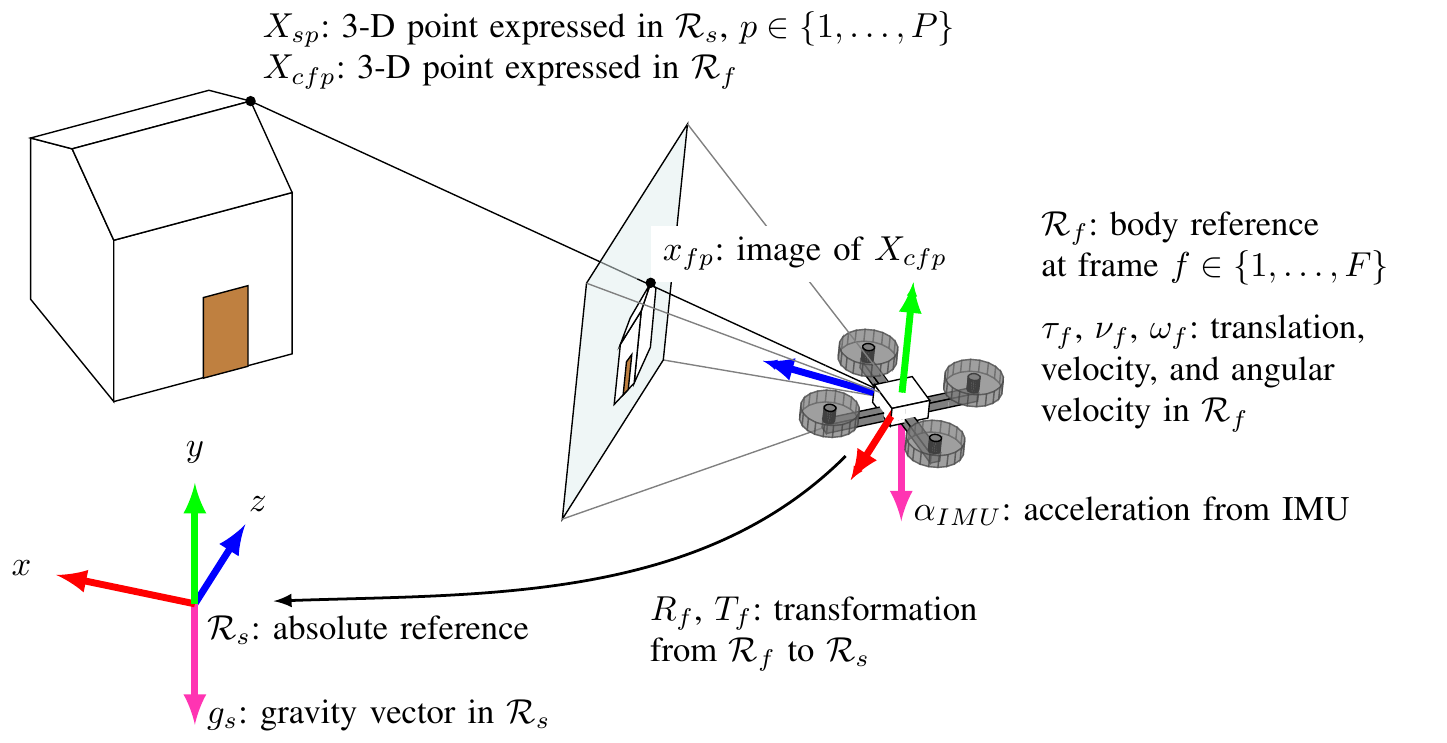}
  \caption{Schematic illustration of the problem and notation.}
  \label{fig:notation}
\end{figure*}

On the other end of the spectrum we have pure \emph{filtering} approaches.
While some approaches are based on the Unscented Kalman Filter (UKF) \cite{Huster:OCEANS02,Ebcin:GNSS01,Kelly:IJRR11} or Particle filter \cite{Fox:JAIR99,Pupilli:BMVC05}, the majority are based on the Extended Kalman Filter (EKF). 
The inertial measurements can be used in either a \emph{loosely coupled} manner, i.e., in the update step of the filter \cite{Konolige:RR11,Oskiper:CVPR07,Chai:PTVE02,Jones:IJRR11}, 
or in a \emph{tightly coupled} manner, i.e., in the prediction step of the filter together with a kinematic model \cite{Jung:CVPR01,Ma:ICRA12,Tardif:IROS10,Brockers:SPIE12,Kottas:RSS13,Strelow:IJRR04,Kim:ICRA03,Kleinert:MFI10,Pinies:ICRA07}. 
Methods based on the EKF can be combined with an inverse depth parametrization \cite{Kleinert:MFI10,Pinies:ICRA07,Civera:RSS06,Eade:CVPR06} to reduce linearization errors.

Between batch and filtering approaches there are three options. The first is to use incremental solutions to the batch problem \cite{Kaess:IJRR11}. 
The second option is to use a \emph{Sliding Window Filter} (SWF) approach, which applies a batch algorithm on a small set of recent measurements. 
The states that are removed from the window are compressed into a prior term using linearization and marginalization 
\cite{Mei:IJCV11,Sibley:JFR10,Leutenegger:IJRR15,DongSi:ICRA11,Huang:IROS11}, possibly approximating the sparsity of the original problem \cite{Nerurkar:ICRA14}.
The third option is to use a \emph{Multistate-Constrained Kalman Filter} (MSCKF), which is similar to a sliding window filter, but where the old states are \emph{stochastic clones} \cite{Roumeliotis:ICRA02} that remain constant and are not updated with the measurements. 
Comparisons of the two approaches \cite{Leutenegger:IJRR15,Clement:CRV15} show that the SWF is more accurate and robust, but the MSCKF is more efficient. 
A hybrid method that switches between the two has appeared in \cite{Li:IROS12}, \cite{Li:RSS13}.

%% file: notation.tex
\section{Notation and preliminaries}
In this section we establish the notation for the following sections. In particular, we define quantities related to a robot (e.g., a quadrotor) equipped with a camera and an inertial measurement unit. We consider its motion as a rigid body, and the relation of this with the IMU measurements and with the geometry of the scene.

\paragraph{Reference frames, transformations and velocities} We first define an inertial spatial frame $\cR_s$, which corresponds to a fixed ``world'' reference frame, and a camera reference frame $\cR_f$, which corresponds to body-fixed reference frame of the robot. For simplicity, we assume that the reference frame of the camera and of the IMU coincide with $\cR_f$, and that they are both centered at the center of mass of the robot. 
We denote the world-centered frame as $\cR_s$, and as $\cR_f$ the robot-centered frame at some time instant (or ``frame'') $f\in\{1,\ldots,F\}$. We also define the pair $(R_f,T_f)\in SE(3)$, where $R_f$ is a 3-D rotation belonging to the space of rotations $SO(3)$, and $T_f\in\real{3}$ is a 3-D translation and $SE(3)$ is the group of rigid body motions \cite{Murray:book94}. 
More concretely, given a point with 3-D coordinates $X_f\in\real{3}$ in the camera frame, the same point in the spatial frame will have coordinates $X_s\in\real{3}$ given by:
\begin{equation}
\label{eq:transformationbs}
  X_s=R_{f}X_c+T_{f}.
\end{equation}
Note that this equation implies that the $T_{f}$ is equal to the position of the center of mass of the vehicle in $\cR_s$. Hence, $\dot{T}_{f}$ and $\ddot{T}_{f}$ represent its velocity and acceleration in the same reference frame.

We also define the angular velocity $\omega_f\in\real{3}$ with respect to $\cR_f$ such that
\begin{equation}
  \dot{R}_f=R_f\hat{\omega}_f,\label{eq:omegab}
\end{equation}

With this notation, Euler's equation of motion for the vehicle can be written as \cite{Murray:book94}:
\begin{equation}
  J\dot{\omega}_f+\hat{\omega}_fJ\omega_f=\Gamma_f, \label{eq:Jdomegac}
\end{equation}
where $J$ is the moment of inertia matrix and $\Gamma_f$ is the torque applied to the body, both defined with respect to the local reference frame $\cR_f$. 

\paragraph{Body-fixed quantities}
We denote the translation, linear velocity, linear acceleration, rotation and angular velocity of the robot expressed in the reference $\cR_f$ as $\tau_f$, $\nu_f$, $\alpha_f$, $R_f$ and $\omega_f$, respectively. Since these are vectors, they are related to the corresponding quantities in the inertial frame $\cR_s$ by the rotation $R_f\transpose$:
\begin{align}
  \tau_f&=R_f\transpose T_f,  \label{eq:taub} \\
  \nu_f&=R_f\transpose\dot{T}_f, \label{eq:nub}\\
  \alpha_f&=R_f\transpose\ddot{T}_f. \label{eq:taub-nub-alphab}
\end{align}

An ideal body-fixed ideal IMU unit will measure the angular velocity
\begin{equation}
  \omega_{IMU}=\omega_f,  
\end{equation}
and the acceleration
A body-fixed, ideal accelerometer positioned at the center of mass of the object will measure
\begin{equation}
  \alpha_{IMU}=R_{f}\transpose(\ddot{T}_{f}+g_s)=\alpha_f+R_{f}\transpose g_s,\label{eq:alphaimu}
\end{equation}
where $g_s$ is the (downward pointing) gravity vector in the spatial frame $\cR_s$, around $-9.8e_z \mathrm{m}/\mathrm{s}^2$, where and $e_z=[0\; 0\; 1]\transpose$.

\paragraph{Tridimensional structure}
We assume that the onboard camera can track the position of $P$ points having coordinates $X_{sp}\in\real{3}$, $p\in\{1,\ldots,P\}$ in $\cR_s$. For convenience, we assume that $\cR_s$ is centered at the centroid of this points, that is, $\frac{1}{P}\sum_{p=1}^PX_{sp}=0$.

Given the quantities above, we can find expressions for the coordinates of a point in the camera's coordinate system and its derivatives. However, it is first convenient to find the derivatives of $\tau_c$, $\nu_c$ and $\omega_c$, which can be obtained by combining \eqref{eq:taub} and \eqref{eq:nub} with the definition \eqref{eq:omegab}, and from Euler's equation of motion \eqref{eq:Jdomegac}.
\begin{align}
  \dot{\tau}_f&=-\hat{\omega}_f\tau_f+\nu_f, \label{eq:dtaub}\\
  \dot{\nu}_f&=-\hat{\omega}_f\nu_f+\alpha_f, \label{eq:dnub}\\
  \dot{\omega}_f&=J\inverse\bigl(\Gamma_f-\hat{\omega}_fJ\omega_f\bigr).\label{eq:domegab}
\end{align}

Then, the coordinate of a point $X_{sp}$ in the reference $\cR_f$ and its derivatives are given by \cite{Murray:book94}:
\begin{align}
  X_{cfp}=&R_f\transpose X_{sp}-\tau_f, \qquad \label{eq:Xc}\\
  \dot{X}_{cfp}=&-\hat{\omega}_f R_f\transpose X_{sp}+\hat{\omega}_f\tau_f-\nu_f, \label{eq:dXc}\\
  \ddot{X}_{cfp}=&\bigl(\hat{\omega}_f^2-\dot{\hat{\omega}}_f) R_f\transpose X_{sp} - \bigl(\hat{\omega}_f^2-\dot{\hat{\omega}}_f) \tau_f\nonumber\\&+2\hat{\omega}_f\nu_f-\alpha^{IMU}_f-R_f\transpose g_s, \label{eq:Xc-dXc-ddXc}
\end{align}
where $\hat{\cdot}$ denotes the skew-symmetric matrix representation of the cross product \cite{Ma:book04}, and where $\dot{\omega}_f$ can be obtained either using Euler's equation of motion as in \eqref{eq:domegab}, by assuming $\dot{\omega}_f=0$ (constant velocity model) or by using numerical differentiation of $\omega_f$.

Note that $X_{cfp}$ and its derivatives these quantities can be completely determined by the 3-D geometry of the scene in the inertial reference frame, the motion of the camera ($R_{sc}$, $\tau_c$, $\nu_c$) and the measurements of the IMU ($\alpha_{IMU}$, $\omega_{IMU}$); these all contain some coefficient matrix times the term $R_f\transpose X_{sp}$ plus a vector given by the IMU measurements, the translational motion of the robot and the gravity vector. This structure will lead to the low-rank factorization formulation below.

\paragraph{Image projections}
The coordinates in the image of the projection of $X_{sp}$ at frame $f$ is denoted as $x_{fp}$. Assuming that the camera is intrinsically calibrated \cite{Ma:book04}, the image $x_{fp}$ can be related to $X_{cfp}$ with the \emph{affine camera model}, that is:
\begin{equation}
  x_{fp}=\Pi X_{cfp}. \label{eq:affine-model}
\end{equation}
where $\Pi\in\real{2 \times 3}$ is a projector that removes the third element of a vector.
This model is an approximation of the projective model for when the scene is relatively far from the camera. This model has been used for \emph{Affine SfM} \cite{Tomasi:IJCV92} and \emph{Affine Motion Segmentation} (see the review article \cite{Vidal:SPM11} and references within), and it will allow us to introduce the basic principles of our proposed methods.

Using \eqref{eq:affine-model}, one can show that the images $\{x_{fp}\}$ and their derivatives $\{\dot{x}_{fp}\}$ (\emph{flow}) and $\{\ddot{x}_{fp}\}$ (\emph{double flow}) can be written as:
\begin{align}
  x_{fp}=&\Pi R_f\transpose X_{sp}-\tau_f, \qquad \label{eq:Xc}\\
  \dot{x}_{fp}=&-\Pi\hat{\omega}_f R_f\transpose X_{sp}+\Pi(\hat{\omega}_f\tau_f-\nu_f), \label{eq:dXc}\\
  \ddot{x}_{fp}=&\Pi\bigl(\hat{\omega}_f^2-\dot{\hat{\omega}}_f\bigr) R_f\transpose X_{sp} - \Pi\bigl((\hat{\omega}_f^2-\dot{\hat{\omega}}_f) \tau_f\\&+2\hat{\omega}_f\nu_f-\alpha^{IMU}_f-R_f\transpose g_s\bigr). \label{eq:ddXc}
\end{align}

\paragraph{Formal problem statement}
 In this section we give the technical details for the proposed Dynamic SfM estimation methods for single agents. The setup and notation used in this section are shown in. We assume that the camera on the robot can track $P$ points for $F$ frames. We assume that the derivatives of the tracked points are available (e.g., through numerical differentiation).
 
Using the notation introduced in this section, the Dynamic Affine SfM problem is then formulated as finding the motion $\{R_f,\tau_f,\nu_f\}$, the structure $\{X_{sp}\}$ and the gravity vector $g_s$ from the camera measurements $\{x_{fp}\},\{\dot{x}_{fp}\},\{\ddot{x}_{fp}\}$ and the IMU measurements $\{\omega_f,\alpha^{IMU}_f\}$. Figure~\ref{fig:notation} contains a graphical summary of the problem and of the notation.



%% file: solution.tex
\section{Dynamic Affine SfM}
 \label{sc:solution}
\paragraph{Factorization formulation}
We start our treatment by collecting all the image measurements and their derivatives in a single matrix
\begin{equation}
  W=\stack(W',\dot{W}',\ddot{W}') \in\real{6F\times P},
\end{equation}
where the matrix $W'\in\real{3F\times P}$ is defined by stacking the coordinates $\{x_{fp}\}$ following the frame index $f$ across the rows and the point index $p$ across the columns:
\begin{equation}
  W'=\bmat{x_{11} & \cdots & x_{1P}\\ \vdots  & \ddots & \vdots \\  x_{N1} & \cdots & x_{NP}} \in\real{2F\times P}.
\end{equation}

Notice the common structure in where we have some coefficient matrix times $R_f\transpose$ times $X_{sp}$ plus a vector. Thus, the matrix $W$ admits an affine rank-three decomposition (which can also be written as a rank four decomposition)
\begin{equation}
W=CMS+m=\bmat{CM & m}\bmat{S\\\vct{1}\transpose},\label{eq:factorization}
\end{equation}
where the \emph{motion matrix}
\begin{equation}
M=\stack\bigl(\{R_f\transpose\}\bigr)\label{eq:M}
\end{equation}
contains the rotations, the \emph{structure matrix} 
\begin{equation}
S=\bmat{X_{s1}&\cdots&X_{sP}}
\end{equation}
contains the 3-D points expressed in $\cR_s$, the \emph{coefficient matrix} $C$ contains the projector $\Pi$ times the coefficients multiplying the rotations in \eqref{eq:Xc-dXc-ddXc}
\begin{equation}
C=\stack\left(\{\Pi\}_{f=1}^F,\{-\Pi\hat{\omega}_{f}\}_{f=1}^F,\{\Pi(\hat{\omega}_f^2-\dot{\hat{\omega}}_f)\}_{f=1}^F\right),
\end{equation}
 and the \emph{translation vector} $m\in\real{2F}$ contains the remaining vector terms
\begin{multline}
m=\stack\bigl(\{-\Pi\tau_{f}\}_{f=1}^F,\{\Pi(\hat{\omega}_f\tau_f-\nu_{f})\}_{f=1}^F,\\
\{\Pi((\hat{\omega}_f^2-\dot{\hat{\omega}}_f) \tau_f +2\hat{\omega}_{f}\nu_{f}-\alpha_{IMU}-R_{f}\transpose g_s)\}_{f=1}^F\bigr).\label{eq:m}
 \end{multline}

In addition to this relation, the quantities $\tau_f$, $\nu_f$ and $\alpha_f^{IMU}$ can be linearly related using derivatives (see \eqref{eq:taub-nub-alphab}). Similarly, $R_f$ and $w_f$ can be related using the definition of angular velocity.
Note that the coefficients $C$ are completely determined by the IMU measurements and the torque inputs.

\paragraph{Optimization formulation}
From this, the problem of estimating the motion, the structure, and the gravity direction can then be casted as an optimization problem:
\begin{multline}
\min_{\{R_f,\tau_f,\nu_f\},g_s}\frob{W-(CMS+m)}^2+f_R(\{R_f\},\{\omega_f\})\\+f_\tau(M,\{\tau_f\},\{\nu_f\})+f_\nu(M,\{\nu_f\},\{\alpha_f^{IMU}\},g_s),  \label{eq:factorization-affine}
\end{multline}
where $f_R$, $f_\tau$ and $f_\nu$ are quadratic regularization terms based on approximating the linear derivative constraints between $\tau_f$, $\nu_f$, $\alpha_f^{IMU}$ and between $R_f$, $w_f$ with finite differences.

In particular, for our implementation we will use,:
\begin{align}
f_R&=\sum_{f=1}^{F-1}\frob{R_{f+1}-R_f\expm(t_s\omega_f)}^2 \label{eq:regR}\\
f_\tau&=\sum_{f=1}^{F-1} \frob{\frac{1}{t_s}\conv(\tau_k,h_k,f)-\nu_f}^2\label{eq:regtau}\\
f_\nu&=\sum_{f=1}^{F-1} \frob{\frac{1}{t_s}\conv(\nu_k,h_k,f)-\alpha_{IMU}-R_f\transpose g_s}^2.\label{eq:regnu}
\end{align}
where with $t_s$ is the sampling period of the measurements, $\expm$ is the matrix exponential and $\conv(\tau_k,h_k,f)$ gives the sample at time $f$ of the convolution $\tau_k\ast h_k$ of a signal $\tau_k$ with a derivative interpolation filter $h_k$. For our implementation we obtain $h_k$ from a Savitzky-Golay filter of order one and window size three.

\paragraph{Solution strategy} The optimization problem \eqref{eq:factorization-affine} is non-convex. However, we can find a closed-form solution by exploiting the low-rank nature of the product $MS$ and the linearity of the other terms. This closed-form solution is exact for the noiseless case, and provides an approximated solution to \eqref{eq:factorization-affine} in the noisy case. 

\begin{enumerate}
\item\label{it:factorization} Factorization. Compute a rank four factorization $W=\tM\tS$ using an SVD. With respect to the last term in \eqref{eq:factorization}, the factors $\tM$ and $\tS$ are related to, respectively $\bmat{CM & m}$ and $\bmat{S\\\vct{1}\transpose}$ by an unknown matrix $\Kproj\in\real{4\times 4}$. In standard SfM terminology, $\tM$ and $\tS$ represent a \emph{projective} reconstruction.
\item Similarity transformation. Ideally, the last row of $\tS'$ should be $\vct{1}\transpose$. Therefore, we find a vector $k\in\real{4}$ by solving $k\transpose \tS'=\vct{1}\transpose$ in a least squares sense. We then define the matrix $\Ksym=\stack(\bmat{I & 0},k\transpose)$, and the matrices $\tM'=\tM \Ksym\inverse$, $\tS'=\Ksym\tS$. In standard SfM terminology, $\tM'$ and $\tS'$ represent reconstruction up to a \emph{symilarity} transformation.
\item\label{it:centering} Centering. To fix the center of the absolute reference frame $\cR_s$ to the center of the 3-D structure, we first compute the vector $c=\frac{1}{P}[\tS']_{1:3,:}$, where $[\tS']_{1:3,:}$ indicates the matrix composed of the first three rows of $\tS'$. We then define the matrix $\Kcent=\bmat{0 & -c\\0 & 1}$, and the matrices $\tM''=\tM' \Kcent\inverse$ and $\tS''=\Kcent\tS'$. At this point the forth column of the matrix $\tM''$ contains (in the ideal case) the vector $m$ defined in \eqref{eq:factorization}, that is $\hat{m}=[\tM'']_{:,4}$.
\item\label{it:rotations} Recovery of  the rotations and structure. We now solve for the rotations $\{R_f\}$ by solving a reduced version of \eqref{eq:factorization-affine}. In particular, we solve
  \begin{equation}
    \min_{M''\in\real{3F\times 3}} \norm{[\tM'']_{:,1:3}-CM''}_F^2+\norm{C_RM''}_F^2,\label{eq:opt-rotation}
  \end{equation}
where $[\tM'']_{:,1:3}$ indicates the matrix containing the first three columns of $\tM''$, and $C_R$ is a block-banded-diagonal matrix with blocks $I$ and $-\expm(t_s\omega_f)\transpose$ corresponding to the regularization term \eqref{eq:regR}. This is a simple least squares problem which can be easily solved using standard linear algebra algorithms. Ideally, the matrix $M''$ is related to the real matrix $M$ by an unknown similarity transformation $\Kupg\in\real{3\times 3}$. This matrix can be determined (up to an arbitrary rotation) using the standard metric upgrade step from Affine SfM (see \cite{Tomasi:IJCV92}). Once $\Kupg$ has been determined, we define $\hat{M}=M''\Kupg$ and $\hat{S}=\Kupg\inverse [\tS'']_{1:3,:}$ to be the estimated motion and structure matrices. The final estimates $\{\hat{R}_f\}$ are obtained by projecting each $3\times 3$ block of $\hat{M}$ to $SO(3)$ using an SVD decomposition.
\item\label{it:translations} Recovery of the translations and linear velocities. We need to extract $\{\tau_f\}$ and $\{\omega_f\}$ and an estimated gravity direction $\hat{g}_s$ from the vector $\hat{m}$. Following~\eqref{eq:m}, we define the matrix
  \begin{equation}
    C_m=\bmat{ \vdots & \vdots & \vdots \\ -\Pi & 0 & 0\\ \vdots & \vdots & \vdots \\ \Pi\hat{\omega}_f & -\Pi & 0\\   \vdots & \vdots & \vdots \\\Pi((\hat{\omega}_f^2-\dot{\hat{\omega}}_f) & 2\hat{\omega}_{f} & -R_{f}\\ \vdots & \vdots & \vdots }
  \end{equation}
and the vector $c_m=\stack(\vct{0}_{6F},\{\alpha_f^{IMU}\})$. Similarly to the definition of $C_R$ in \eqref{eq:opt-rotation}, we also define the matrices $C_\tau$, $C_\nu$ corresponding to the regularization terms \eqref{eq:regtau} and \eqref{eq:regnu}.
We can then solve for the vector $x=\stack(\{\hat{\tau}_f\},\{\hat{\nu_f\}},\hat{g}_s)$ by minimizing
\begin{equation}
  \min_x \norm{C_mx-\hat{m}}_F^2+\norm{C_\tau x}_F^2+\norm{C_\nu x}_F^2,\label{eq:opt-translation}
\end{equation}
which again is a least squares problem that can be solved using standard linear algebra tools.
\end{enumerate}

%% file: 2016-cvpr-dynamicSfM.bbl
\begin{thebibliography}{10}\itemsep=-1pt

\bibitem{Aftab:PAMI15}
K.~Aftab, R.~Hartley, and J.~Trumpf.
\newblock Generalized weiszfeld algorithms for lq optimization.
\newblock {\em {IEEE} Transactions on Pattern Analysis and Machine
  Intelligence}, 37(4):728--745, 2015.

\bibitem{Agarwal:COMM11}
S.~Agarwal, Y.~Furukawa, N.~Snavely, I.~Simon, B.~Curless, S.~M. Seitz, and
  R.~Szeliski.
\newblock Building {R}ome in a day.
\newblock {\em Communications of the {ACM}}, 54(10):105--112, 2011.

\bibitem{Agarwal:ECCV10}
S.~Agarwal, N.~Snavely, S.~M. Seitz, and R.~Szeliski.
\newblock Bundle adjustment in the large.
\newblock In {\em {IEEE} European Conference on Computer Vision}, pages 29--42.
  Springer, 2010.

\bibitem{Agrawal:IROS06}
M.~Agrawal.
\newblock A lie algebraic approach for consistent pose registration for general
  euclidean motion.
\newblock In {\em {IEEE} International Conference on Intelligent Robots and
  Systems}, pages 1891--1897, 2006.

\bibitem{Arie-Nachimson:3DIMPVT12}
M.~Arie-Nachimson, S.~Kovalsky, I.~Kemelmacher-Shlizerman, A.~Singer, and
  R.~Basri.
\newblock Global motion estimation from point matches.
\newblock In {\em International Conference on 3D Imaging, Modeling, Processing,
  Visualization and Transmission}, pages 81--88, 2012.

\bibitem{Bay:CVIU08}
H.~Bay, A.~Ess, T.~Tuytelaars, and L.~V. Gool.
\newblock Speeded-up robust features {(SURF)}.
\newblock {\em Computer Vision and Image Understanding}, 110(3):346--359, 2008.

\bibitem{Brockers:SPIE12}
R.~Brockers, S.~Susca, D.~Zhu, and L.~Matthies.
\newblock Fully self-contained vision-aided navigation and landing of a micro
  air vehicle independent from external sensor inputs.
\newblock In {\em SPIE Defense, Security, and Sensing}, page 83870Q.
  International Society for Optics and Photonics, 2012.

\bibitem{Bryson:ICRA09}
M.~Bryson, M.~Johnson-Roberson, and S.~Sukkarieh.
\newblock Airborne smoothing and mapping using vision and inertial sensors.
\newblock In {\em {IEEE} International Conference on Robotics and Automation},
  pages 2037--2042, 2009.

\bibitem{Carlone:ICRA15}
L.~Carlone, R.~Tron, K.~Daniilidis, and F.~Dellaert.
\newblock Initialization techniques for {3D SLAM}: a survey on rotation
  estimation and its use in pose graph optimization.
\newblock In {\em {IEEE} International Conference on Robotics and Automation},
  2015.

\bibitem{Chai:PTVE02}
L.~Chai, W.~A. Hoff, and T.~Vincent.
\newblock Three-dimensional motion and structure estimation using inertial
  sensors and computer vision for augmented reality.
\newblock {\em Presence: Teleoperators and Virtual Environments},
  11(5):474--492, 2002.

\bibitem{Chatterjee:ICCV13}
A.~Chatterjee and V.~M. Govindu.
\newblock Efficient and robust large-scale rotation averaging.
\newblock In {\em {IEEE} International Conference on Computer Vision}, pages
  521--528, 2013.

\bibitem{Civera:RSS06}
J.~Civera, A.~J. Davison, and J.~M.~M. Montiel.
\newblock Unified inverse depth parametrization for monocular slam.
\newblock In {\em Robotics: Science and Systems}, 2006.

\bibitem{Clement:CRV15}
L.~E. Clement, V.~Peretroukhin, J.~Lambert, and J.~Kelly.
\newblock The battle for filter supremacy: A comparative study of the
  multi-state constraint {K}alman {F}ilter and the {S}liding {W}indow {F}ilter.
\newblock In {\em {IEEE} Conference on Computer and Robot Vision}, pages
  23--30, 2015.

\bibitem{Nerurkar:ICRA14}
E.~D.~N. D, K.~J. Wu, and S.~Roumeliotis.
\newblock {C-KLAM}: Constrained keyframe-based localization and mapping.
\newblock In {\em {IEEE} International Conference on Robotics and Automation},
  pages 3638--3643, 2014.

\bibitem{Dai:IJCV13}
Y.~Dai, H.~Li, and M.~He.
\newblock Projective multiview structure and motion from element-wise
  factorization.
\newblock {\em {IEEE} Transactions on Pattern Analysis and Machine
  Intelligence}, 35(9):2238--2251, 2013.

\bibitem{Dellaert:08}
F.~Dellaert and et~al.
\newblock Georgia tech smoothing and mapping ({GTSAM}).
\newblock \url{http://tinyurl.com/gtsam}.

\bibitem{Dong:CVPR15}
J.~Dong and S.~Soatto.
\newblock Domain-size pooling in local descriptors: {DSP-SIFT}.
\newblock In {\em {IEEE} Conference on Computer Vision and Pattern
  Recognition}, pages 5097--5106, 2015.

\bibitem{DongSi:ICRA11}
T.-C. Dong-Si and A.~I. Mourikis.
\newblock Motion tracking with fixed-lag smoothing: Algorithm and consistency
  analysis.
\newblock In {\em {IEEE} International Conference on Robotics and Automation},
  pages 5655--5662, 2011.

\bibitem{DongSi:IROS12}
T.-C. Dong-Si and A.~I. Mourikis.
\newblock Estimator initialization in vision-aided inertial navigation with
  unknown camera-imu calibration.
\newblock In {\em {IEEE} International Conference on Intelligent Robots and
  Systems}, pages 1064--1071, 2012.

\bibitem{Eade:CVPR06}
E.~Eade and T.~Drummond.
\newblock Scalable monocular {SLAM}.
\newblock In {\em {IEEE} Conference on Computer Vision and Pattern
  Recognition}, volume~1, pages 469--476, 2006.

\bibitem{Ebcin:GNSS01}
S.~Ebcin and M.~Veth.
\newblock Tightly-coupled image-aided inertial navigation using the unscented
  kalman filter.
\newblock In {\em International Technical Meeting of the Satellite Division of
  The Institute of Navigation (GNSS)}, pages 1851--1860, 2001.

\bibitem{Engels:PCV06}
E.~C. Engels, H.~Stew{\'e}nius, and D.~Nist{\'e}r.
\newblock Bundle adjustment rules.
\newblock In {\em Photogrammetric Computer Vision}, volume~2, pages 124--131,
  2006.

\bibitem{Fischler:CommACM81}
M.~A. Fischler and R.~C. Bolles.
\newblock Random sample consensus: a paradigm for model fitting with
  applications to image analysis and automated cartography.
\newblock {\em Communications of the {ACM}}, 24(6):381--395, 1981.

\bibitem{Fox:JAIR99}
D.~Fox, W.~Burgard, and S.~Thrun.
\newblock Markov localization for mobile robots in dynamic environments.
\newblock {\em Journal of Artificial Intelligence Research}, pages 391--427,
  1999.

\bibitem{Frahm:ECCV10}
J.-M. Frahm, P.~Fite-Georgel, D.~Gallup, T.~Johnson, R.~Raguram, C.~Wu, Y.-H.
  Jen, E.~Dunn, B.~Clipp, S.~Lazebnik, and M.~Pollefeys.
\newblock Building {R}ome on a cloudless day.
\newblock In {\em {IEEE} European Conference on Computer Vision}, pages
  368--381. Springer, 2010.

\bibitem{Hartley:pami12}
R.~Hartley and H.~Li.
\newblock An efficient hidden variable approach to minimal-case camera motion
  estimation.
\newblock {\em {IEEE} Transactions on Pattern Analysis and Machine
  Intelligence}, 2012.

\bibitem{Hartley:IJCV13}
R.~Hartley, J.~Trumpf, Y.~Dai, and H.~Li.
\newblock Rotation averaging.
\newblock {\em International Journal of Computer Vision}, 103(3):267--305,
  2013.

\bibitem{Hartley:book04}
R.~I. Hartley and A.~Zisserman.
\newblock {\em Multiple View Geometry in Computer Vision}.
\newblock Cambridge University Press, second edition, 2004.

\bibitem{Huang:IROS11}
G.~P. Huang, A.~I. Mourikis, and S.~Roumeliotis.
\newblock An observability-constrained sliding window filter for slam.
\newblock In {\em {IEEE} International Conference on Intelligent Robots and
  Systems}, pages 65--72, 2011.

\bibitem{Huster:OCEANS02}
A.~Huster, E.~W. Frew, and S.~M. Rock.
\newblock Relative position estimation for auvs by fusing bearing and inertial
  rate sensor measurements.
\newblock In {\em MTS/IEEE OCEANS}, volume~3, pages 1863--1870, 2002.

\bibitem{Indelman:NDRV15}
V.~Indelman and F.~Dellaert.
\newblock Incremental light bundle adjustment: Probabilistic analysis and
  application to robotic navigation.
\newblock In {\em New Development in Robot Vision}, pages 111--136. Springer,
  2015.

\bibitem{Jones:WDV07}
E.~Jones, A.~Vedaldi, and S.~Soatto.
\newblock Inertial structure from motion with autocalibration.
\newblock In {\em Workshop on Dynamical Vision}, 2007.

\bibitem{Jones:IJRR11}
E.~S. Jones and S.~Soatto.
\newblock Visual-inertial navigation, mapping and localization: A scalable
  real-time causal approach.
\newblock {\em The International Journal of Robotics Research}, 30(4):407--430,
  2011.

\bibitem{Jung:CVPR01}
S.-H. Jung and C.~J. Taylor.
\newblock Camera trajectory estimation using inertial sensor measurements and
  structure from motion results.
\newblock In {\em {IEEE} Conference on Computer Vision and Pattern
  Recognition}, volume~2, pages --732. IEEE, 2001.

\bibitem{Kaess:IJRR11}
M.~Kaess, H.~Johannsson, R.~Roberts, V.~Ila, J.~J. Leonard, and F.~Dellaert.
\newblock {iSAM2}: Incremental smoothing and mapping using the bayes tree.
\newblock {\em The International Journal of Robotics Research}, pages 1--20,
  2011.

\bibitem{Kelly:IJRR11}
J.~Kelly and G.~S. Sukhatme.
\newblock Visual-inertial sensor fusion: Localization, mapping and
  sensor-to-sensor self-calibration.
\newblock {\em The International Journal of Robotics Research}, 30(1):56--79,
  2011.

\bibitem{Kim:ICRA03}
J.-H. Kim and S.~Sukkarieh.
\newblock Airborne simultaneous localisation and map building.
\newblock In {\em {IEEE} International Conference on Robotics and Automation},
  volume~1, pages 406--411, 2003.

\bibitem{Kleinert:MFI10}
M.~Kleinert and S.~Schleith.
\newblock Inertial aided monocular {SLAM} for {GPS}-denied navigation.
\newblock In {\em {IEEE} Conference on Multisensor Fusion and Integration for
  Intelligent Systems}, pages 20--25, 2010.

\bibitem{Konolige:TRO08}
K.~Konolige and M.~Agrawal.
\newblock {FrameSLAM}: From bundle adjustment to real-time visual mapping.
\newblock {\em {IEEE} Transactions on Robotics and Automation},
  24(5):1066--1077, 2008.

\bibitem{Konolige:RR11}
K.~Konolige, M.~Agrawal, and J.~Sola.
\newblock Large-scale visual odometry for rough terrain.
\newblock In {\em Robotics Research}, pages 201--212. Springer, 2011.

\bibitem{Kottas:RSS13}
D.~G. Kottas and S.~I. Roumeliotis.
\newblock Exploiting urban scenes for vision-aided inertial navigation.
\newblock In {\em Robotics: Science and Systems}, 2013.

\bibitem{Leutenegger:IJRR15}
S.~Leutenegger, S.~Lynen, M.~Bosse, R.~Siegwart, and P.~Furgale.
\newblock Keyframe-based visual-inertial odometry using nonlinear optimization.
\newblock {\em The International Journal of Robotics Research}, 34(3):314--334,
  2015.

\bibitem{Li:IROS12}
M.~Li and A.~Mourikis.
\newblock Vision-aided inertial navigation for resource-constrained systems.
\newblock In {\em {IEEE} International Conference on Intelligent Robots and
  Systems}, pages 1057--1063, 2012.

\bibitem{Li:RSS13}
M.~Li and A.~I. Mourikis.
\newblock Optimization-based estimator design for vision-aided inertial
  navigation.
\newblock In {\em Robotics: Science and Systems}, pages 241--248, 2013.

\bibitem{Lobo:IJRR07}
J.~Lobo and J.~Dias.
\newblock Relative pose calibration between visual and inertial sensors.
\newblock {\em The International Journal of Robotics Research}, 26(6):561--575,
  2007.

\bibitem{Lowe:IJCV04}
D.~G. Lowe.
\newblock Distinctive image features from scale-invariant keypoints.
\newblock {\em International Journal of Computer Vision}, 60(2):91--110, 2004.

\bibitem{Lupton:TRO12}
T.~Lupton and S.~Sukkarieh.
\newblock Visual-inertial-aided navigation for high-dynamic motion in built
  environments without initial conditions.
\newblock {\em {IEEE} Transactions on Robotics and Automation}, 28(1):61--76,
  2012.

\bibitem{Lynen:IROS13}
S.~Lynen, M.~W. Achtelik, S.~Weiss, M.~Chli, and R.~Siegwart.
\newblock A robust and modular multi-sensor fusion approach applied to {MAV}
  navigation.
\newblock In {\em {IEEE} International Conference on Intelligent Robots and
  Systems}, pages 3923--3929, 2013.

\bibitem{Ma:ICRA12}
J.~Ma, S.~Susca, M.~Bajracharya, L.~Matthies, M.~Malchano, and D.~Wooden.
\newblock Robust multi-sensor, day/night {6-DOF} pose estimation for a dynamic
  legged vehicle in {GPS}-denied environments.
\newblock In {\em {IEEE} International Conference on Robotics and Automation},
  pages 619--626, 2012.

\bibitem{Ma:book04}
Y.~Ma.
\newblock {\em An invitation to 3-{D} vision: from images to geometric models}.
\newblock Springer, 2004.

\bibitem{Martinec:CVPR07}
D.~Martinec and T.~Pajdla.
\newblock Robust rotation and translation estimation in multiview
  reconstruction.
\newblock In {\em {IEEE} Conference on Computer Vision and Pattern
  Recognition}, pages 1--8, 2007.

\bibitem{Martinelli:TRO12}
A.~Martinelli.
\newblock Vision and imu data fusion: Closed-form solutions for attitude,
  speed, absolute scale, and bias determination.
\newblock {\em {IEEE} Transactions on Robotics and Automation}, 28(1):44--60,
  2012.

\bibitem{Martinelli:FTROB13}
A.~Martinelli.
\newblock Observabilty properties and deterministic algorithms in
  visual-inertial structure from motion.
\newblock {\em Foundations and Trends in Robotics}, pages 1--75, 2013.

\bibitem{Mei:IJCV11}
C.~Mei, G.~Sibley, M.~Cummins, P.~Newman, and I.~Reid.
\newblock {RSLAM}: A system for large-scale mapping in constant-time using
  stereo.
\newblock {\em International Journal of Computer Vision}, 94(2):198--214, 2011.

\bibitem{Murray:book94}
R.~M. Murray, Z.~Li, and S.~S. Sastry.
\newblock {\em A Mathematical Introduction to Robotic Manipulation}.
\newblock CRC Press, 1994.

\bibitem{Oskiper:CVPR07}
T.~Oskiper, Z.~Zhu, S.~Samarasekera, and R.~Kumar.
\newblock Visual odometry system using multiple stereo cameras and inertial
  measurement unit.
\newblock In {\em {IEEE} Conference on Computer Vision and Pattern
  Recognition}, pages 1--8, 2007.

\bibitem{Pinies:ICRA07}
P.~Pini{\'e}s, T.~Lupton, S.~Sukkarieh, and J.~D. Tard{\'o}s.
\newblock Inertial aiding of inverse depth slam using a monocular camera.
\newblock In {\em {IEEE} International Conference on Robotics and Automation},
  pages 2797--2802, 2007.

\bibitem{Pupilli:BMVC05}
M.~Pupilli and A.~Calway.
\newblock Real-time camera tracking using a particle filter.
\newblock In {\em British Machine Vision Conference}, 2005.

\bibitem{Roumeliotis:ICRA02}
S.~Roumeliotis, A.~E. Johnson, and J.~F. Montgomery.
\newblock Augmenting inertial navigation with image-based motion estimation.
\newblock In {\em {IEEE} International Conference on Robotics and Automation},
  volume~4, pages 4326--4333. IEEE, 2002.

\bibitem{Sibley:JFR10}
G.~Sibley, L.~Matthies, and G.~Sukhatme.
\newblock Sliding window filter with application to planetary landing.
\newblock {\em Journal of Field Robotics}, 27(5):587--608, 2010.

\bibitem{Snavely:TOG06}
N.~Snavely, S.~M. Seitz, and R.~Szeliski.
\newblock Photo tourism: exploring photo collections in {3D}.
\newblock In {\em {ACM} Transactions on Graphics}, volume~25, pages 835--846,
  2006.

\bibitem{Snavely:CVPR08}
N.~Snavely, S.~M. Seitz, and R.~Szeliski.
\newblock Skeletal graphs for efficient structure from motion.
\newblock In {\em {IEEE} Conference on Computer Vision and Pattern
  Recognition}, volume~1, page~2, 2008.

\bibitem{Strelow:IJRR04}
D.~Strelow and S.~Singh.
\newblock Motion estimation from image and inertial measurements.
\newblock {\em The International Journal of Robotics Research},
  23(12):1157--1195, 2004.

\bibitem{Tardif:IROS10}
J.-P. Tardif, M.~George, M.~Laverne, A.~Kelly, and A.~Stentz.
\newblock A new approach to vision-aided inertial navigation.
\newblock In {\em {IEEE} International Conference on Intelligent Robots and
  Systems}, pages 4161--4168, 2010.

\bibitem{Tomasi:IJCV92}
C.~Tomasi and T.~Kanade.
\newblock Shape and motion from image streams under orthography: a
  factorization method.
\newblock {\em International Journal of Computer Vision}, 9(2):137--154, 1992.

\bibitem{Triggs:VATP00}
B.~Triggs, P.~F. McLauchlan, R.~I. Hartley, and A.~W. Fitzgibbon.
\newblock Bundle adjustment -- a modern synthesis.
\newblock In {\em Vision algorithms: theory and practice}, pages 298--372.
  Springer, 2000.

\bibitem{Tron:TAC14}
R.~Tron and R.~Vidal.
\newblock Distributed {3-D} localization of camera sensor networks from {2-D}
  image measurements.
\newblock {\em {IEEE} Transactions on Automatic Control}, 2014.

\bibitem{Vidal:SPM11}
R.~Vidal.
\newblock Subspace clustering.
\newblock {\em {IEEE} Signal Processing Magazine}, 2(28):52--68, 2011.

\bibitem{Wang:II13}
L.~Wang and A.~Singer.
\newblock Exact and stable recovery of rotations for robust synchronization.
\newblock {\em Information and Inference}, 2(2):145--193, 2013.

\bibitem{Weiss:ICRA12}
S.~Weiss, M.~W. Achtelik, S.~Lynen, M.~Chli, and R.~Siegwart.
\newblock Real-time onboard visual-inertial state estimation and
  self-calibration of {MAV}s in unknown environments.
\newblock In {\em {IEEE} International Conference on Robotics and Automation},
  pages 957--964, 2012.

\end{thebibliography}
